\documentclass[runningheads]{llncs}

\usepackage{eccv}

\usepackage{eccvabbrv}

\usepackage{graphicx}
\usepackage{booktabs}

\usepackage[accsupp]{axessibility}  % Improves PDF readability for those with disabilities.

\usepackage{hyperref}

\usepackage{orcidlink}

\begin{document}

\title{LASPA: Latent Spatial Alignment for Fast Training-free Single Image Editing} 

\titlerunning{LASPA}

\author{Yazeed Alharbi\inst{1}\orcidlink{0000-0002-8073-1959} \and
Peter Wonka\inst{2}\orcidlink{0000-0003-0627-9746}}

%\authorrunning{F.~Author et al.}

\institute{NCAI \and KAUST}

\maketitle
\begin{center}
    \centering
    \captionsetup{type=figure}
    \includegraphics[width=1\textwidth]{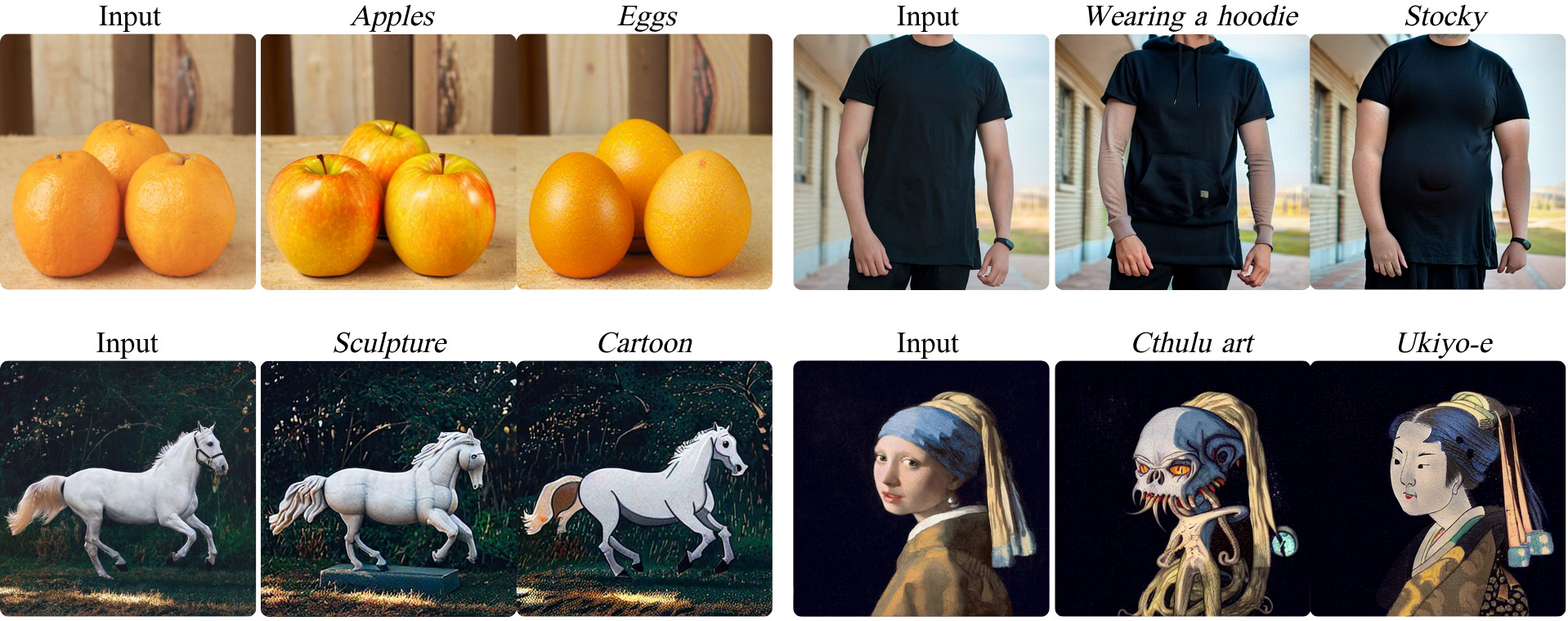}
    \captionof{figure}{Using a single real image as input, our method is capable of editing using textual prompts in less than 6 seconds without finetuning the diffusion model or using costly image embedding algorithms. Our results show accurate editing for realistic as well as artistic edits.}
\end{center}

\begin{abstract}
We present a novel, training-free approach for textual editing of real images using diffusion models. Unlike prior methods that rely on computationally expensive finetuning, our approach leverages LAtent SPatial Alignment (LASPA) to efficiently preserve image details. We demonstrate how the diffusion process is amenable to spatial guidance using a reference image, leading to semantically coherent edits. This eliminates the need for complex optimization and costly model finetuning, resulting in significantly faster editing compared to previous methods. Additionally, our method avoids the storage requirements associated with large finetuned models. These advantages make our approach particularly well-suited for editing on mobile devices and applications demanding rapid response times. While simple and fast, our method achieves 62-71\% preference in a user-study and significantly better model-based editing strength and image preservation scores.
\end{abstract}

\section{Introduction}
\label{sec:intro}
Text-to-image diffusion models~\cite{rombach2022highresolution, saharia2022imagen, latentdiffusion, ramesh2022dalle2, ding2021cogview, balaji2022eDiff-I, dalle3} have revolutionized image generation, creating realistic and diverse images from textual prompts. However, harnessing these models for single-image editing remains a challenge, as it requires preserving the original image details while generating the required edits.\newline
Traditionally, editing with generative models involves inverting the input image into the latent space of the generator to preserve details, and then manipulating the latent to achieve the desired edit. Diffusion models offer advantages for both inversion and editing due to their spatial latents. The latents carry rich spatial information and can be easily reproduced given real images using the noise schedule.\par
Despite the potential of the spatial latents, state-of-the-art single-image editing methods~\cite{kawar2022imagic, SINE, nulltext} rely on textual tokens while finetuning the model’s parameters to encode reference image information. This forgoes the advantages of spatial latents, leading to longer processing times and increased storage needs.
 While the last two years have brought fantastic improvements in diffusion-based image editing, the computational cost and memory overhead of these methods are significant. Existing work~\cite{SINE, kawar2022imagic, nulltext, dreambooth, gal2022image} is confined to editing speeds in the order of minutes while requiring the storage of a separate finetuned model per image or edit. However, image-editing applications require results almost instantly, as users on mobile devices, desktop applications, or cloud-based services, are only willing to wait for a few seconds to achieve an edit. Further, cloud-based services require cheap computation costs to be profitable, and mobile devices require simple and memory-efficient computation in general. \par 

To address these limitations, we do not propose acceleration strategies of existing methods, but an alternative method on how edits can be achieved that is conceptually much simpler. We present LASPA, a novel approach for single-image editing using text-to-image diffusion models. We propose leveraging real image information to guide the spatial latents, while keeping the textual latents intact. By eliminating the need for optimization and finetuning, LASPA addresses the computational constraints that may otherwise limit the widespread adoption of diffusion-based image editing. \par
Specifically, in lieu of optimizing or finetuning to preserve the input image, we align any random latent using reference image features directly using different parts of the reverse diffusion process. Our knowledge of the amount of noise added to training samples at each timestep enables aligning the input sample $x_t$. Our knowledge of the denoising UNet allows aligning the predicted error $\epsilon_{\theta}$. Our knowledge of a good initial estimate for $x_0$ enables alignment of the prediction of $x_0$. \par
We perform qualitative and quantitative experiments to validate the gains of our method. We show that our method is consistently preferred by users, and leads to better image preservation and editing strength model-based scores. %Our method, while simple and fast, leads consistently to higher quality of editing when measured by human-based and model-based metrics.
%Specifically, our contributions include three distinct approaches to latent spatial alignment, accurate extraction of object attention maps based on aligned latents, and a background improvement method using semantic latent mixing. Our experimental results demonstrate a significant enhancement in image editing performance, achieving orders of magnitude faster processing compared to fine-tuning methods while maintaining superior quality results in terms of input image preservation and editing strength.

%The results show that our contributions lead to a significant improvement in image editing performance while requiring only inference steps to edit. In comparison to methods that require finetuning~\cite{kawar2022imagic, SINE, nulltext}, our method is orders of magnitudes faster while requiring only the storage of the pretrained stable-diffusion weights. In comparison with faster methods~\cite{SDEDIT, couairon2022diffedit}, our method produces significantly higher quality results in terms of input image preservation and editing strength.

\section{Related work}
%\textbf{Diffusion models for unconditional image generation:}

\textbf{Diffusion models for image generation:}
Diffusion models~\cite{ho2020denoising, song2021ddim, latentdiffusion, ditxl, dhariwal2021diffusion, wang2023patchdiffusion, zheng2022truncateddiffusion, ho2022cascadeddiffusion} offer excellent results for image synthesis. %To train the generative models, real images are diffused by adding noise progressively. The generative part, which is the reverse diffusion process, is tasked with removing the noise and recreating the real images. Specifically, for stable-diffusion~\cite{rombach2022highresolution} a transformer UNet is trained to achieve this task. 
%Recent unconditional diffusion models~\cite{latentdiffusion, ditxl, dhariwal2021diffusion, wang2023patchdiffusion, zheng2022truncateddiffusion, ho2022cascadeddiffusion} achieve competitive results on several unconditional generation datasets.\newline
A unique aspect of diffusion models is that they operate on images or spatial latents extracted by pretrained VQ models~\cite{esser2021taming}. This is a major deviation from previous generative models, GANs, where vectors were used as latents.\newline For editing with unconditional diffusion models DiffusionCLIP~\cite{kim2022diffusionclip} shows interesting results. Given a diffusion model trained on a specific class (for example, faces), the authors show how the CLIP loss can be used to perform textual edits. Crucially, DiffusionCLIP shows that image reconstruction is noticeably more accurate with diffusion models than GANs. FreeDoM~\cite{yu2023freedom} proposes training-free conditional editing for unconditional models, such that images or text can be used.\newline
Recent diffusion models~\cite{ding2021cogview, latentdiffusion, rombach2022highresolution, ramesh2022dalle2,balaji2022eDiff-I, dalle3, pernias2023wuerstchen, sdxl} offer prompt-based control over the generated images. They allow almost unlimited control over the generated images such as generating in a certain artist's style or generating fictional creatures. This is achieved through cross-attention between a spatial image latent and a textual latent. Given the amount of training data and the quality of results, a new area of research aspired to leverage the learned text-image associations for other tasks.\newline
\textbf{Controllable generation with text-to-image diffusion models:}
One of the earlier works for controlling images generated by diffusion models is prompt-to-prompt~\cite{prompt2prompt}. Prompt-to-prompt shows that attention maps learned by stable diffusion are useful to preserve the semantic layout of generated images. %Therefore, to edit a generated image while maintaining the layout, prompt-to-prompt proposes swapping the attention maps of the edited images with the original generated image. The intuition is to use cross-attention during editing to remain similar to the original image. 
However, prompt-to-prompt fails to preserve the identities of objects. Therefore it requires specialized embedding algorithms, such as Null-text inversion~\cite{nulltext} to edit real images. %Therefore, prompt-to-prompt cannot be used on real images. Furthermore, it cannot be used to edit a generated image using another generated image with a different seed.
One strand of work is to train separate auxiliary networks to inject conditioning information, such as segmentation or depth maps. Currently, ControlNet~\cite{zhang2023adding} and Lora~\cite{hu2022lora} are very popular for this task. \newline
\textbf{Real image editing with text-to-image diffusion models:}
Several works in the literature propose editing improvements such as better inversion methods~\cite{pan2023effective, wallace2023edict}, better scoring functions~\cite{hertz2023delta}, and improvements to editing with stochastic models~\cite{wu2023latent}.
More recently, there are two emerging research problems: model personalization and single image editing.\newline
Model personalization aims to capture a concept using a few images such that the text-to-image model can regenerate it in novel views and edit it. While single image editing uses text-to-image models to enable text-based editing of real images given only a single image.
Most previous methods for single image editing or model personalization~\cite{SINE, dreambooth, kawar2022imagic, gal2022image} utilize the textual embedding to encode the identity of objects to be preserved. As a result, previous methods require either finetuning the entire model or optimizing the textual token. These methods require minutes to edit while requiring the storage of a finetuned model per edit or image.\newline
While some prior methods enable faster editing~\cite{SDEDIT, blended, couairon2022diffedit, brooks2023instructpix2pixdiffusion}, they are very limited in comparison with methods that require finetuning. SDEdit~\cite{SDEDIT} often fails to preserve identity information, while blended diffusion~\cite{blended} and DiffEdit~\cite{couairon2022diffedit} are restricted to local edits. InstructPix2Pix~\cite{brooks2023instructpix2pixdiffusion} offers inference time editing using textual instructions and shows convincing results. However, it requires training a model with a large generated set of text and image pairs.
\newcommand{\enci}{enc(I)}

\section{Method}
\label{sec:method}

\subsection{Motivation}
To motivate our approach, we review diffusion model training and inference.
Diffusion models are trained by adding noise to training images according to a predefined schedule. This is referred to as the diffusion process. The noised training sample $x_t$ is a linear combination of original image information $x_0$ and added noise $\epsilon \sim \mathcal{N}({0}, \textit{{Id}})$ according to predetermined weights $\alpha$:
\begin{equation}
x^t = \sqrt{\alpha_t}x_0 + \sqrt{1-\alpha_t}\epsilon
\end{equation}
A crucial observation is that the reference image in editing tasks can be easily diffused to any timestep. This can be done by adding noise according to the training schedule. \par%A crucial observation is that any real image can be easily diffused to any timestep by blending according to the schedule.\newline
The transformer UNet is trained to predict the noise/error at any timestep through the following loss function:
\begin{equation}
\mathcal{L}(\epsilon_{\theta}) = \lVert \epsilon_{\theta}^{(t)} ( \sqrt{\alpha_t}x_0 + \sqrt{1-\alpha_t}\epsilon_t) - \epsilon_t \rVert ^{2} _{2}
\end{equation}
Another important observation is that we can also easily compute the reconstruction error at any timestep of the reverse diffusion process. This can be done by computing the difference between the reference image and the partially denoised sample at the current timestep.\par %Another important observation is that the error between any real image and any sample can be easily computed for any timestep by computing their difference. \newline
We use DDIM sampling~\cite{song2021ddim}, where the error $\epsilon_{\theta}^{(t)}$ predicted by the transformer UNet is used to predict the final result $x_0$ from the current result $x_t$:
\begin{equation}
pred_{x0}^{(t)} = \frac{x_t - \sqrt{1-\alpha_t}\epsilon_{\theta}^{(t)} (x_t)}{\sqrt{\alpha_t}}
\end{equation}
This prediction of $x_0$ is used to compute the next denoised estimation $x_{t-1}$ given the current $x_t$:
\begin{equation}
x_{t-1} = \sqrt{\alpha_{t-1}} pred_{x0}^{(t)} + \sqrt{1-\alpha_{t-1}} \epsilon_{\theta}^{(t)} (x_t) \label{eq:ddim}
\end{equation}
Where the right-hand term of the addition is referred to as the direction pointing to $x_t$. We observe that for the task of editing, an initial prediction of $x_0$ can be easily obtained directly using the reference image.\par
Our observations about the generative reverse diffusion process form the basis of our methodology. We claim that several components of the generative reverse diffusion process are receptive to spatial guidance using reference image features directly.

\begin{figure}[t]
  \centering
 \includegraphics[width=0.8\textwidth]{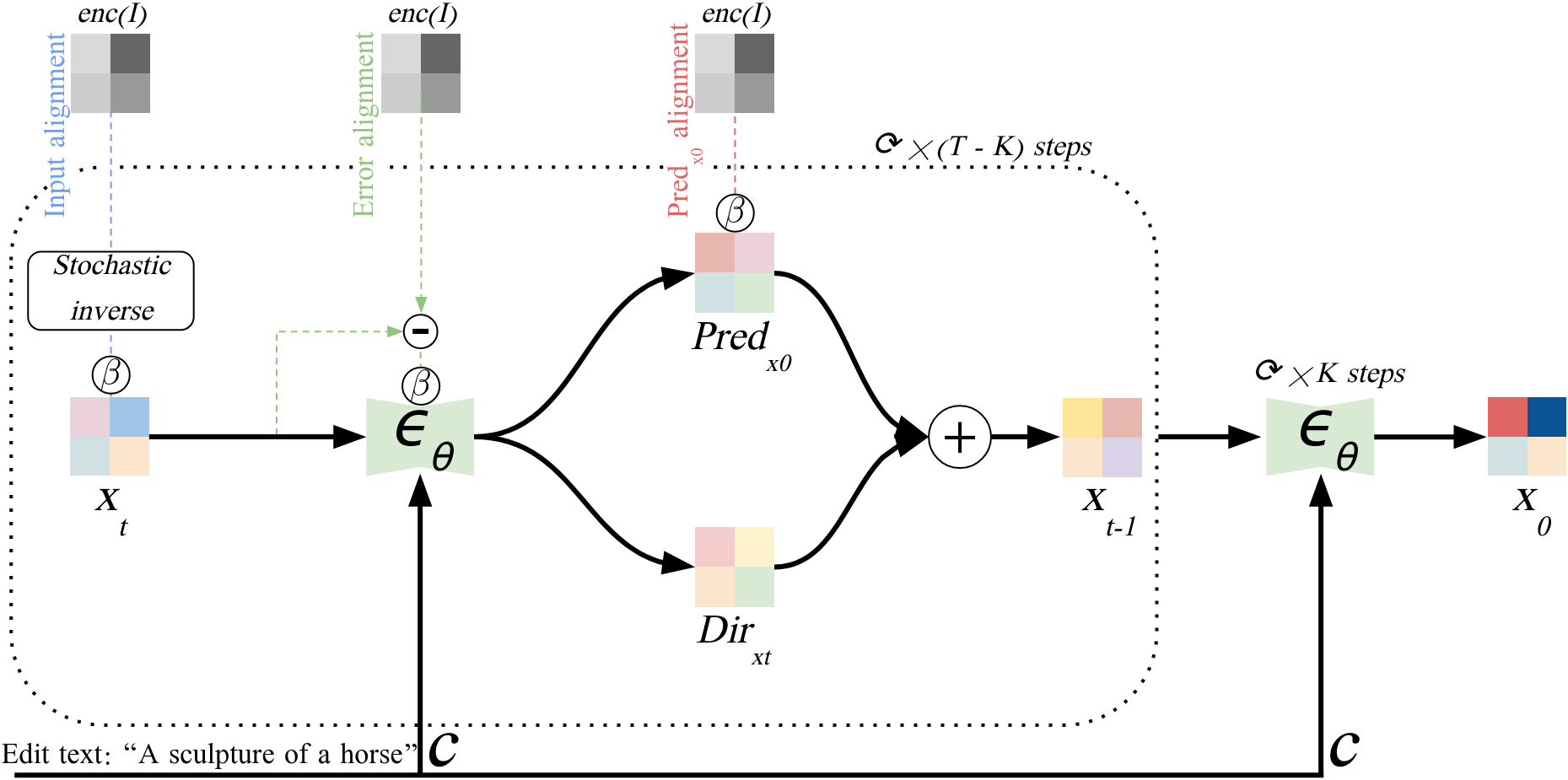}
  \caption{An illustration of our alignment methods and how they influence reverse diffusion steps.}
\label{fig:methof}
\end{figure}

\subsection{Latent spatial alignment}
Given a reference image $I$ and a prompt $P$, the goal of prompt-based image editing is to produce an image $\hat{I}$ that remains as similar as possible to $I$ while reflecting the edits specific by $P$.\par
There are two types of latents in text-to-image diffusion models: spatial latents and textual latents. The spatial latents are noised visual information being either images $I$ or latents $\enci$ extracted using a pre-trained autoencoder (e.g. the $64\times64\times4$ tensor that is gradually transformed in Stable Diffusion~\cite{rombach2022highresolution}). The textual latents are extracted from $P$ and are used to condition the generation results. Contrary to the common approach of encoding the image into textual latents, we propose latent spatial alignment to edit using reference image features directly, without requiring optimization or changing the model's parameters.  \par
LASPA demonstrates how different components of the reverse diffusion process are receptive to latent spatial alignment. The key insight is that we can align the network using real image information to produce a reverse diffusion path that starts by aiming to reconstruct $I$ while slowly introducing the edits highlighted in $P$.\newline
All of our proposed alignment paths follow this intuition. They are training-free, they align only the spatial latents, and they lead to \textit{exact} reconstruction of $enc(I)$ if the reconstruction path is followed for all reverse diffusion steps. However, they differ significantly in how they guide reconstruction and how they can be used to guide editing.
One reason for our achieved speedup is that the proposed alignment strategies eliminate the need for image inversion as we can initialize with arbitrary noise $x_T$.
%In the sections below, we show different approaches of using real image information to guide the reverse diffusion path towards the required edits.

\subsection{Input alignment}
The most direct way to encourage the preservation of input image features is to inject the reference features into a certain timestep. This is possible because we have access to the exact training schedule used to introduce noise at each timestep of diffusion. However, directly injecting a noised version of the input image at a certain timestep is problematic. Injecting early allows for editing, but fails to preserve details of the image. This is the approach used in SDEdit~\cite{SDEDIT}. Injecting later preserves almost all the details of the input image but heavily restricts editing.\par
Instead of a one-step injection, we propose aligning the spatial latent with the input at each timestep using reference image features. We start with an unrestricted random noise input, and we slowly align it to the stochastic inversion at the start of each timestep. The stochastic inverse is a weighted combination of a sampled random noise $z \sim \mathcal{N}(0,Id)$ and the encoded input image $\enci$.\newline% (assuming a latent diffusion model such as Stable Diffusion, where $enc$ is computed by a fixed autoencoder).\newline
This leverages the stochastic inversion to preserve spatial details of the input image while allowing $P$ freedom for editing.\newline 
\begin{equation}
z \sim \mathcal{N}(0,Id),\quad \beta_t  = \frac{t}{T},\quad K \in [0, T]
\end{equation}
\begin{equation}
x_{t}^{stoch inv} = \sqrt{\alpha_t} \times \enci + \sqrt{1 - \alpha_t} \times z
\end{equation}
\begin{equation}
    \hat{x_{t}} = 
\begin{cases}
     \beta_t \times x_{t}^{stoch inv} + (1-\beta_t) \times x_{t},& t > K\\
    x_{t}              & t \leq K
\end{cases}
\end{equation}

\begin{figure}[t]
  \centering
 \includegraphics[width=0.8\textwidth]{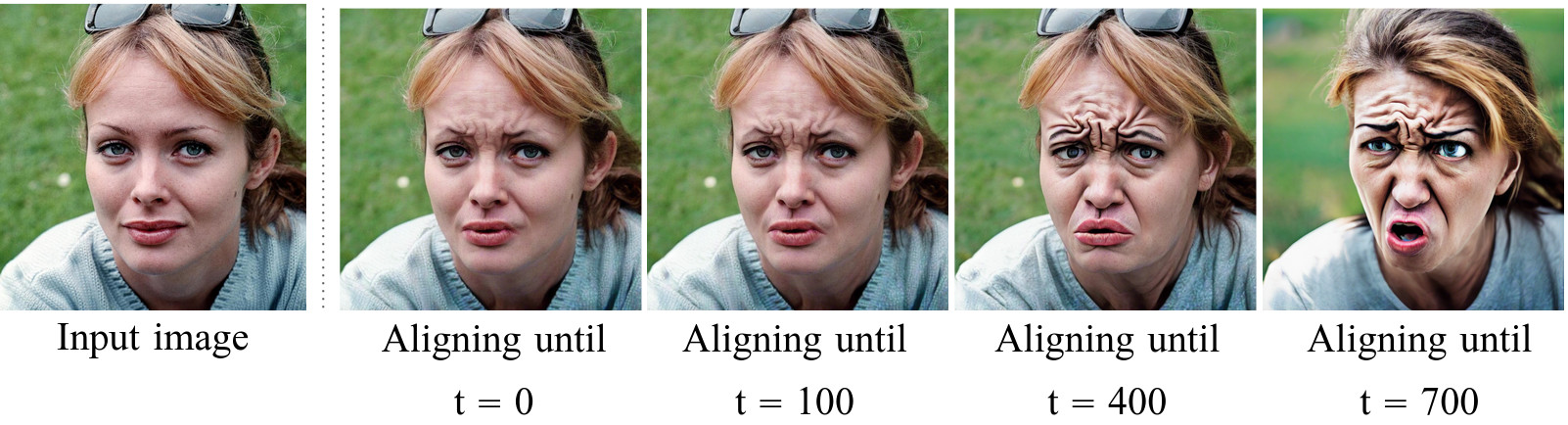}
  \caption{Our latent alignment smoothly incorporates information from both the editing prompt and the input image. The editing prompt used is "a photo of an angry woman."}
    \label{fig:latalign}
\end{figure}

Where $\beta$ controls the alignment strength, $K$ controls the alignment interval, and $\alpha_t$ is a scheduling weight that follows the training procedure.\par
As shown in Figure~\ref{fig:latalign}, we find that the aligned latent begins by encoding large-scale features of the reference image, while reflecting the textual edit. For example, aligning until $t=700$ aligns the pose and color information, aligning until $t=400$ adds the eyeglasses and hairstyle, and aligning until $100$ adds most identity information.\par
There are two interesting aspects about input alignment. First, it aligns using the stochastic inverse. Second, input alignment directly aligns the input $x_t$ to the UNet at each timestep, which can overwrite edits in previous steps that contradict image information. As a result, it leads to high image preservation and low editing strength. %While it is less suitable for generic edits, it is useful for facial editing as shown in the supplementary material.

\subsection{$\epsilon_{\theta}$ alignment}
Instead of directly aligning the input, possibly overwriting some editing details, we can indirectly align samples by aligning the output of the UNet $\epsilon_{\theta}$. While the UNet is trained for unconditional and conditional denoising, we can align it to preserve the input image. Given the latent of the input image $\enci$, we can compute the actual error between $x_t$ and $\enci$ exactly. We use this to align the denoising error to the reconstruction error. This follows the same intuition of input alignment: the aligned latent should first move towards recreating $\enci$, then slowly move towards reflecting $P$.\newline
$\hat{\epsilon}_{\theta}(x_t) = $
\begin{equation} 
\begin{cases}
     \beta_t \times (x_t - \enci) + (1-\beta_t) \times \epsilon_{\theta}(x_t),& t > K\\
     \epsilon_{\theta}(x_t)              & t \leq K
\end{cases}
\end{equation}
Input alignment and error alignment can work with DDPM or DDIM samplers. However, there are important differences. First, error alignment uses the difference with the current sample to align. As a result, it is more dependent on previous steps. Second, in DDIM samplers, error alignment affects both terms of the update: prediction of $x_0$ and direction of $x_t$. This is illustrated in Figure~\ref{fig:methof} and Equation~\ref{eq:ddim}. The computed error $\epsilon_t$ is \textit{subtracted} to obtain the prediction of $x_0$, and then \textit{added} to obtain $x_{t-1}$. Consequently, error alignment is more difficult to tune and leads to higher editing strength at the cost of lower image preservation.

\subsection{Prediction of $x_0$ alignment}
Finally, we propose a more intuitive alignment that follows our reasoning of aligning between the reconstruction path and the editing path. Since DDIM~\cite{song2021ddim} samplers use the error $\epsilon_{\theta}(x_t)$ to predict the denoising result $x_0$ at every step, we can align this prediction using reference image features directly. The intuition is that in early steps the prediction should be close to $\enci$, while in later steps the prediction should be allowed more freedom to introduce $P$.

\begin{equation}
    \hat{pred}_{x0}^{(t)} =
\begin{cases}
     \beta_t \times \enci + (1-\beta_t) \times pred_{x0}^{(t)},& t > K\\
    pred_{x0}^{(t)}             & t \leq K
\end{cases}
\end{equation}
By simply aligning the prediction of $x_0$, any random input can be aligned to reflect both image details while reflecting the textual edit. In comparison with input and error alignment, this alignment method uses the reference image encoding directly to align. Furthermore, this method imposes the least restrictions as it only aligns a part of the DDIM update. As a result, it provides the best tradeoff between editing strength and image preservation, and we use it for all evaluation results in the paper.

\subsection{Real image understanding}
Our approach aims to leverage spatial information of real images without finetuning. This raises the concern of whether our method simply pastes real image information over images conditioned on $P$ leading to artifacts or contradicting information. To adress this concern, we examine the attention maps produced by the UNet during inference. We run our method with prompts that contain the coarse class of the object in the image. For example, ``an image of a \textit{panda}''. Then, we visualize the attention maps of the coarse class word. \par
\begin{figure}[t]
  \centering
 \includegraphics[width=\textwidth]{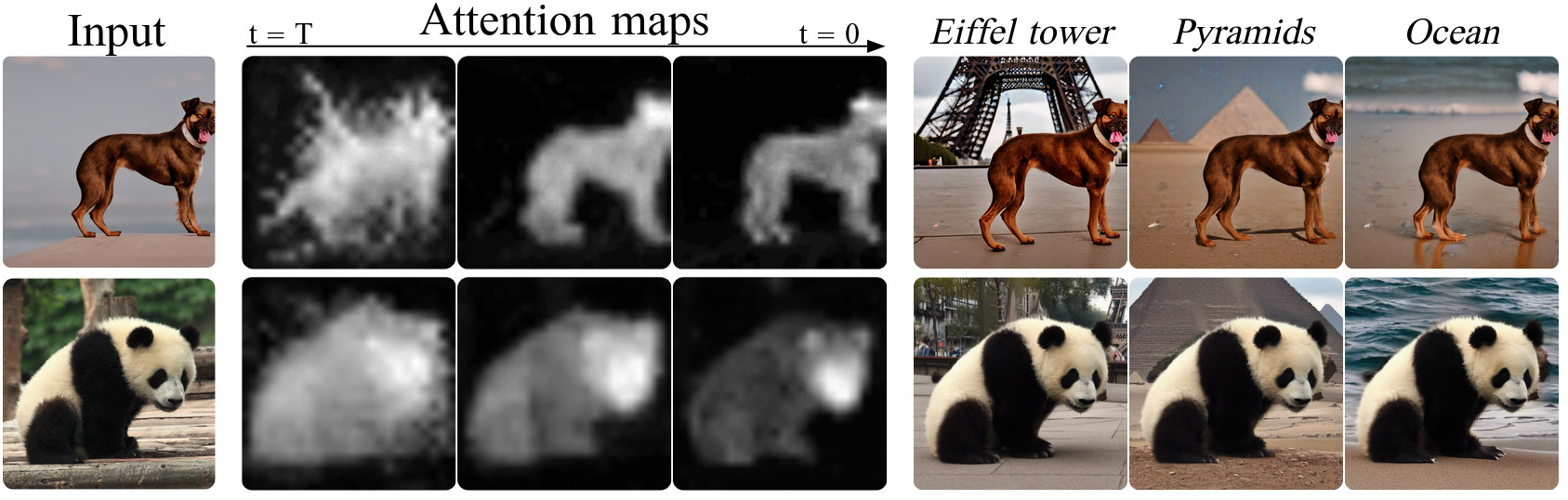}
  \caption{Our proposed alignment methods lead to accurate attention maps and can be used for background editing using semantic latent mixing.}
\label{fig:semanticmixing}

\end{figure}

 As seen in Figure~\ref{fig:semanticmixing}, we find that our proposed alignment leads to accurate attentions maps for real images.
The impact is that our alignment for editing leads to accurate training-free attention maps. In comparison, methods for editing real images based on prompt-to-prompt typically require minutes of optimization to produce attention maps. %In addition, we show that our attention maps are much more accurate than DiffEdit~\cite{couairon2022diffedit} while being much simpler to obtain.

\subsection{Semantic latent mixing}
To further demonstrate the semantic coherence of our method, we leverage the accurate attention maps obtained through alignment to improve editing results in certain cases. In cases where large background changes are needed, we synthesize the final image such that information pertaining to the coarse class will be influenced by the aligned latent, thus maintaining the identity. On the other hand, information pertaining to the rest of the prompt will be influenced by a latent conditioned on $P$.\par
In essence, we use the attention maps of the coarse object class as an editing mask for alignment. Thus, the semantic latent mixing updated would be:

\begin{equation}
    \hat{x}_{t}^{(t)} =
\begin{cases}
     x_t^{aligned} \bullet M + x_t \bullet (1-M),& t > K\\
   x_t             & t \leq K
\end{cases}
\end{equation}

We observe in Figure~\ref{fig:semanticmixing} that the final result of editing is semantically coherent, such that information is exchanged between $I$ and the editing details. This is accomplished through both the self-attention and the cross-attention of the diffusion model. Self-attention, by design, allows each region within the mixed latent to be affected by other regions. We also adjust the cross-attention mechanism to further improve the coherence of the final image. We swap the attention maps of the mixed latent with the attention maps of the aligned latent, similar to prompt-to-prompt. This improves cases where the reverse diffusion process adds another object of the same class in a different location.\newline
We use semantic latent mixing to showcase the semantic coherence of our method in Figures~\ref{fig:semanticmixing} and~\ref{fig:attncomp}. However, for all other results, we use latent alignment without semantic mixing.
%Our semantic latent mixing ensures several aspects in the final image. First, the edited image will follow the general outline of the input image. Second, the coarse class object will be edited without changing the identity. Third, the edited image will strongly reflect the textual edits wherever it does not conflict with the identity of the coarse class object.
\section{Evaluation}
%To showcase the performance of our method, we compare quantitatively and qualitatively against state-of-the-art methods: DiffEdit~\cite{couairon2022diffedit}, SDEdit~\cite{SDEDIT}, Null-text inversion~\cite{nulltext}, SINE~\cite{SINE}, and Imagic~\cite{kawar2022imagic}. We demonstrate the performance in different kinds of edits, such as object replacement and artistic styles. 
\subsection{Implementation details}
For all our qualitative and quantitative experiments, we use stable-diffusion 1.4 with $512\times512$ resolution. However, in the supplementary material, we show additional results on stable-diffusion 2.1 with resolution $768\times768$. For most visual and tabular results in this paper, we choose $K=200$, $\beta=0.3$, the stable-diffusion prompt scale to be $10$, the number of steps to be $50$, and we use $pred_{x_0}$ alignment. %Since the TEdBench contains large pose changes, we set $K=400$ to allow larger freedom for changing image details. We provide further justification for our design choices in the ablations section.\par
Similar to Asyrp~\cite{asyrp}, we find that breaking the symmetry by using a different $\epsilon_{\theta}$ improves editing results.
\begin{figure}[t]
  \centering
 \includegraphics[width=\textwidth]{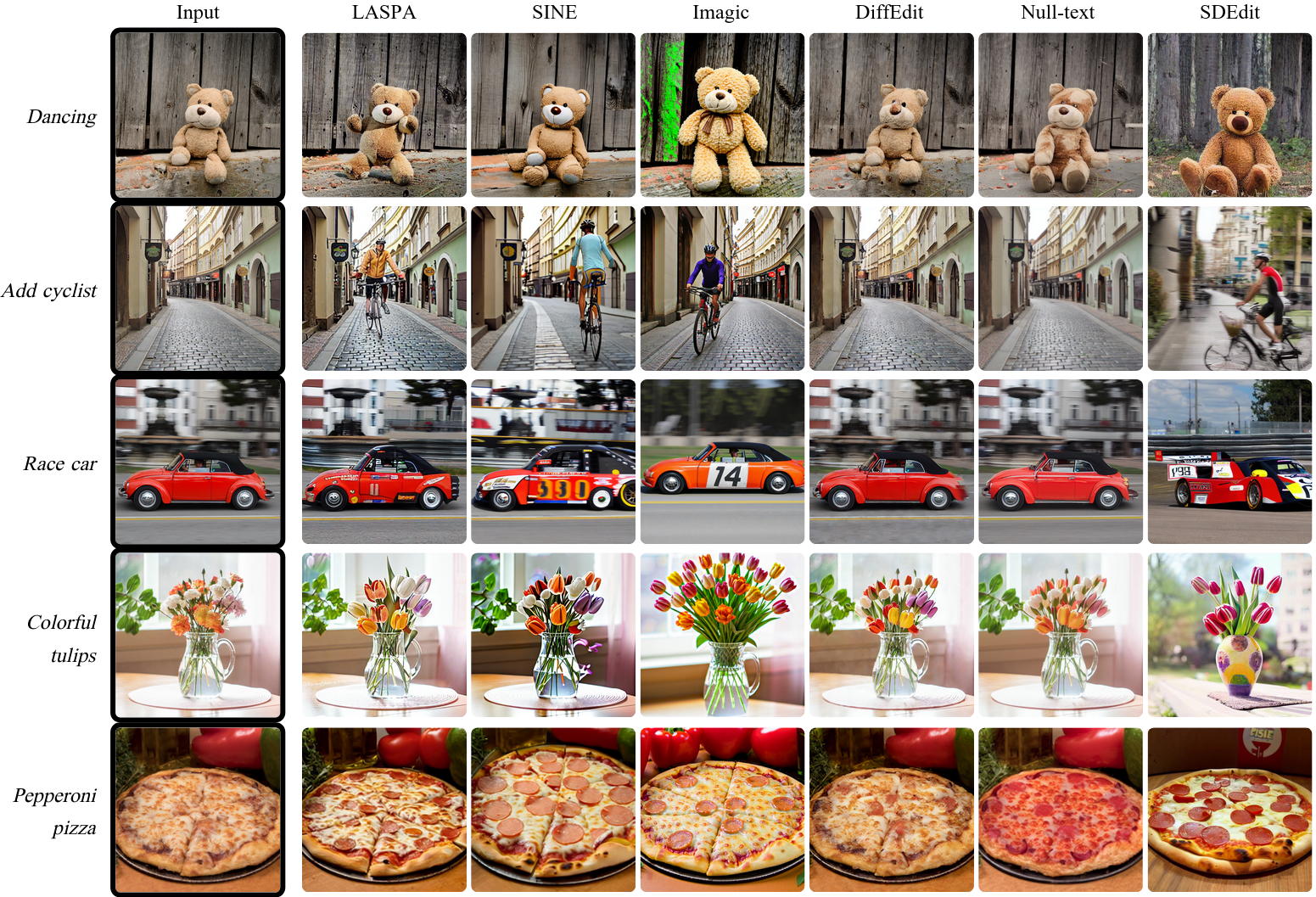} %{figures/fullcomp2.jpg}
  \caption{A visual comparison of editing results between state-of-the-art methods and our method (LASPA). Our method exceeds the quality of methods that require finetuning, while editing only at inference time.}%We can observe our method preserves the foreground object better than SINE, while maintaining the unedited part of the scene better than Imagic. SDEdit produces unsatisfactory results when extended to real image editing. DiffEdit and Null-text inversion excel at object replacement but struggle with other edits.}
\label{fig:qualcomp}
\end{figure}

\begin{figure}[t]
  \centering
 \includegraphics[width=0.45\textwidth]{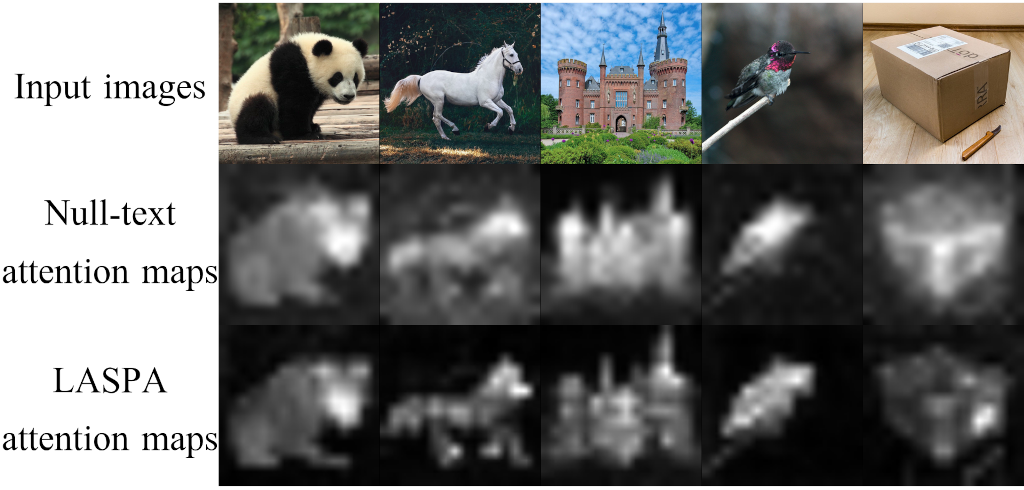}
  \caption{While null-text inversion requires minutes for inversion, LASPA aligns the spatial latent at inference time while producing less noisy attention maps.}
\label{fig:attncomp}
\end{figure}
\subsection{Qualitative comparison}
%We showcase the visual quality of our semantic latent mixing both for real and generated image editing in Figure~\ref{fig:semanticmixing}. Our method leads to better generation of the background while maintaining identity preservation. Although the latents used are the result of attention based mixing of two different latents, the final generated results show semantic coherence. For example, we can see that shadows and reflection are added appropriately for regions outside the object latent.\par
In all explored examples, it is evident that our utilization of the spatial latent to encode image details, rather than textual tokens, leads to higher quality of editing. We compare against SINE~\cite{SINE}, Imagic~\cite{kawar2022imagic}, and Null-text inversion~\cite{nulltext}. Additionally, we include results against faster textual editing methods: DiffEdit~\cite{couairon2022diffedit} and SDEdit~\cite{SDEDIT}. The results can be seen in Figure~\ref{fig:qualcomp}. Against methods that require training (Imagic, SINE, and Null-text inversion) our method leads to much better preservation of input image details while still reflecting the required edits. Against faster methods (SDEdit and DiffEdit) our method is more advantageous in all aspects. DiffEdit and Null-text inversion are more suitable for object replacements and do not introduce most required edits. SDEdit reflects the edits but strays far from input image details. Our method, while being as fast as SDEdit and DiffEdit leads to higher quality editing than methods that require finetuning such as SINE and Imagic.\par
To highlight the semantic coherence of our method, we compare the quality of extract attention maps for different objects. As seen in Figure~\ref{fig:attncomp}, our method leads to less noisy attention maps when compared to Null-text inversion. Null-text inversion requires minutes of optimization to invert the input image. In comparison, our method extracts the attention maps relating to the object keyword during inference-time alignment. The accuracy of the attention maps suggests that reference image features are used to guide the semantics of the final generated image.

%Replaced the smaller table with a user study
\begin{figure}[b]
  \centering
 \includegraphics[width=0.75\textwidth]{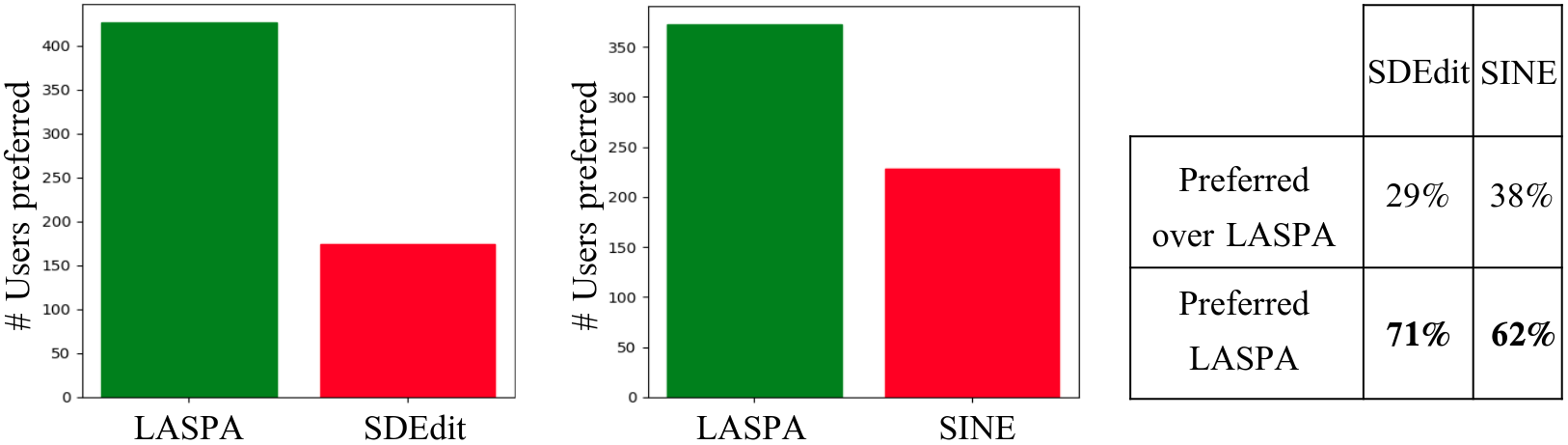}
  \caption{Our method is preferred by users to the inference-time method SDEdit over 70\% of the time. In comparison to SINE, which requires finetuning, our method is preferred over 60\% of the time.}
\label{fig:userstudy}
\end{figure}

\subsection{Quantitative comparison}
We perform two sets of experiments to validate our gains. First, we conduct a user study and ask users to select the editing result they prefer. Second, similar to previous works, we use model-based similarity metrics such as LPIPS~\cite{zhang2018lpips}, and CLIP text similarity~\cite{radford-2021-clip}.\par %We assess the input image preservation using LPIPS and CLIP image embedding similarity between the input image and the output. We assess the influence of the prompt using the CLIP text-image similarity between the prompt and the output.\newline
For the user study, we compare against SDEdit, which offers inference-time editing, and SINE, which requires finetuning. We collect over 1200 responses on different images and edits. As shown in Figure~\ref{fig:userstudy}, our method is preferred significantly more often than SDEdit and SINE. While our method is inference-time, it is still preferable by users to SINE which requires over 15 minutes of finetuning per image.\par
We also make use of the TEdBench dataset~\cite{kawar2022imagic}, which contains 100 edits on many images.  We compare against Imagic, SINE, and SDEdit. As shown in Table~\ref{tab:computedquan2}, our method significantly reduces perceptual distance without harming editing strength. For LASPA, SINE, and SDEdit, we produce multiple results with different tradeoffs to showcase how our method leads to better results.

\begin{table}[t]
\begin{center}
\begin{tabular}{ |c|c c| c c| } 
 \hline
  & CLIP-T $\uparrow$ & LPIPS $\downarrow$ & Time & Additional Storage \\ 
  \hline 
  \textit{Default} &  &  &  & \\
 Imagic & 0.3035  & 0.5925 & 8 (m) & 8GB per edit\\
 SINE  & 0.3035 & 0.5120 & 15 (m) & 8GB per image\\ 
  SDEdit & 0.3031  & 0.5630 & 6 (s) & 0GB  \\ 
 Ours  & \textbf{0.3042} & \textbf{0.4579} & 6 (s) & 0GB \\
 %\textit{(K=400)}  & \textbf{0.3042} & \textbf{0.4579} & ~4 (s) & ~8GB \\
  %\textit{(K=200)} & 0.2918 & \textbf{0.2815} & ~4 (s) & ~8GB\\
 %\hline
 %DiffEdit & 0.2773  & 0.0947 & ~4 (s) & ~8GB \\ 
 
  \hline
  \textit{Config A} &  &  &  & \\
     SINE  & 0.3000 & 0.4110 & 15 (m) &  8GB per image\\ 
  SDEdit & 0.2969  & 0.5560 & 6 (s) &0GB  \\ 
   Ours  & \textbf{0.3001} & \textbf{0.3337} & 6 (s) & 0GB \\
     \hline
  \textit{Config B} &  &  &  & \\
   SINE  & 0.2973 & 0.2817 & 15 (m) & 8GB per image\\ 
  SDEdit & 0.2870  & 0.3094 & 6 (s) & 0GB  \\ 
   Ours  & \textbf{0.2978} & \textbf{0.2501} & 6 (s) & 0GB \\
     \hline

\end{tabular}
\caption{A comparison of image preservation and edit strength similarity scores as measured by CLIP text-image similarity and LPIPS on the TEdBench dataset. Our method leads to the highest editing strength while being the most similar to the input image. Configs A and B explore different preservation-editing tradeoffs.}%Note that DiffEdit is only suitable for object replacement edits.}
\label{tab:computedquan2}
\end{center}
\end{table}

\subsection{Time and memory consumption}
We compare the time and memory consumption of our method against SINE, SDEdit, and Imagic in Table~\ref{tab:computedquan2}. \newline
SINE requires over 15 minutes given the reference image to finetune the network for future edits. Imagic requires around 8 minutes of finetuning for each edit. Our method, on the other hand, can handle any edit in less than 6 seconds. In terms of memory, SINE requires storing a finetuned model per input image, and requires using the outputs of two different networks at inference time. Imagic requires storing a finetuned model and a textual embedding for each pair of reference image and edit prompt. In comparison, our method does not require additional storage or running different models at inference time. We use the same pre-trained stable-diffusion model for all input images and all edit prompts. \newline
While SDEdit is fast and requires storing only a single stable-diffusion model, it leads to significantly worse results both in terms of input preservation and editing strength.

\begin{figure}[t]
  \centering
 \includegraphics[width=\textwidth]{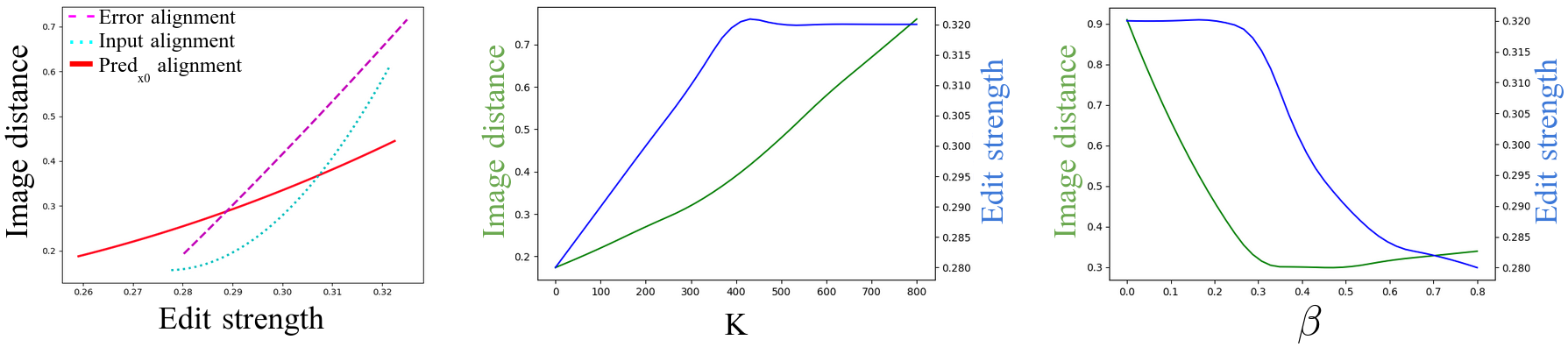}
  \caption{Left: A comparison of the preservation-editing tradeoff between the different alignment methods. Middle and right: plots showing how image preservation and editing strength are affected by parameter choice.}
\label{fig:alignmentab}
\end{figure}
\subsection{Ablation studies}
To perform the ablation studies, we compute the LPIPS image distance and CLIP text similarity metrics on a smaller subset of images and editing prompts. The analysis is shown in Figure~\ref{fig:alignmentab}. \newline First, we analyze the behavior of the various alignment methods in the left subplot of the figure. Input alignment leads to the highest image preservation if alignment is performed in all steps. This is because it directly aligns the input samples, possibly disregarding edits performed in earlier steps. However, for more drastic edits, when the alignment interval and strength are reduced, input alignment struggles with input preservation. We hypothesize that this is due to the usage of the stochastic inverse which is dominated by noise in early steps. This results in lower preservation if alignment is terminated early.\newline%However, we find that input alignment is highly sensitive to $K$ which defines the alignment interval and $\beta$ which defines the alignment strength. When $K$ or $\beta$ are reduced, input alignment can lead to lower image preservation. We hypothesize that this is due to the usage of the stochastic inverse which is dominated by noise in early steps. This results in lower preservation if alignment is terminated early.\newline %Unlike error and prediction of $x_0$ alignment methods, input alignment uses the stochastic inverse to align. In earlier steps, the stochastic inverse is dominated by the noise which is why terminating the alignment early negatively impacts image preservation. Another consequence of the use of stochastic inversion is that input alignment generally requires at least 50 steps of inference to produce good results.\newline
Prediction of $x_0$ leads to the best tradeoff between image preservation and editing strength, especially when larger edits are needed. This is due to how it only aligns one branch of the DDIM update. It uses the input image features directly, and it can generate acceptable results with as low as 20 steps.\newline
Error alignment is the most difficult to tune, as the error appears in opposite signs in the DDIM update. We find that it requires a higher $\beta$ than other methods to preserve the input image details.
\par Additionally, We evaluate the influence of two main parameters: the alignment interval $K$, and the alignment strength $\beta$ to understand their influence on the final performance. As shown in the middle and right subplots in Figure~\ref{fig:alignmentab}, our method works reasonably well on a wide range of parameter choices. The tradeoff between image preservation and editing strength can be controlled through the parameter choices. Intuitively, running our method with a longer alignment interval (lower $K$), and higher alignment strength (higher $\beta$) leads to higher image preservation at the cost of editing strength.\par%For brevity, we include the ablation results on prediction of $x_0$ alignment, and omit ablation results on error and input alignment.\newline
 %The influence of these parameters is visualized in Figure~\ref{fig:ablationvisual}.

\subsection{Discussion}
The qualitative and quantitative analysis consistently show that our method leads to better image preservation than prior methods while still improving faithfulness to editing prompts. The results support our claim that the spatial latent is more suitable for encoding image details than the textual latent.\par
SINE and Imagic finetune stable-diffusion parameters based on a single image. Consequently, some of their editing results show artifacts or saturated colors leading to lower quality and lower image preservation. In addition, they necessitate storing a separate finetuned model per image or per edit, which is not scalable as shown in Table~\ref{tab:computedquan2}. SDEdit and DiffEdit do not require finetuning, but they are more suitable for specific edits. SDEdit performs poorly when extended to textual edits, and is more suitable for conditioning using segmentation maps. DiffEdit excels at localized and object replacement edits. However, it struggles with artistic, global, and object addition edits.\newline
In contrast, our approach aligns entirely within the pretrained spatial and textual latents of text-to-image diffusion models. This leads to several essential benefits. Our method is significantly faster and significantly more similar to the input image, while surpassing the editing strength of previous methods. Furthermore, our method can handle both local and global kinds of edits\par
%Our ablation studies show that our method can be easily adjusted to fit the specific image. Results on edits such as facial editing can be improved by decreasing prompt scale and $K$, and increasing steps and $\beta$ as shown in the supplementary materials. Additionally, we highlight some of the contrast between the three alignment paths.
\begin{figure}[t]
  \centering
 \includegraphics[width=\textwidth]{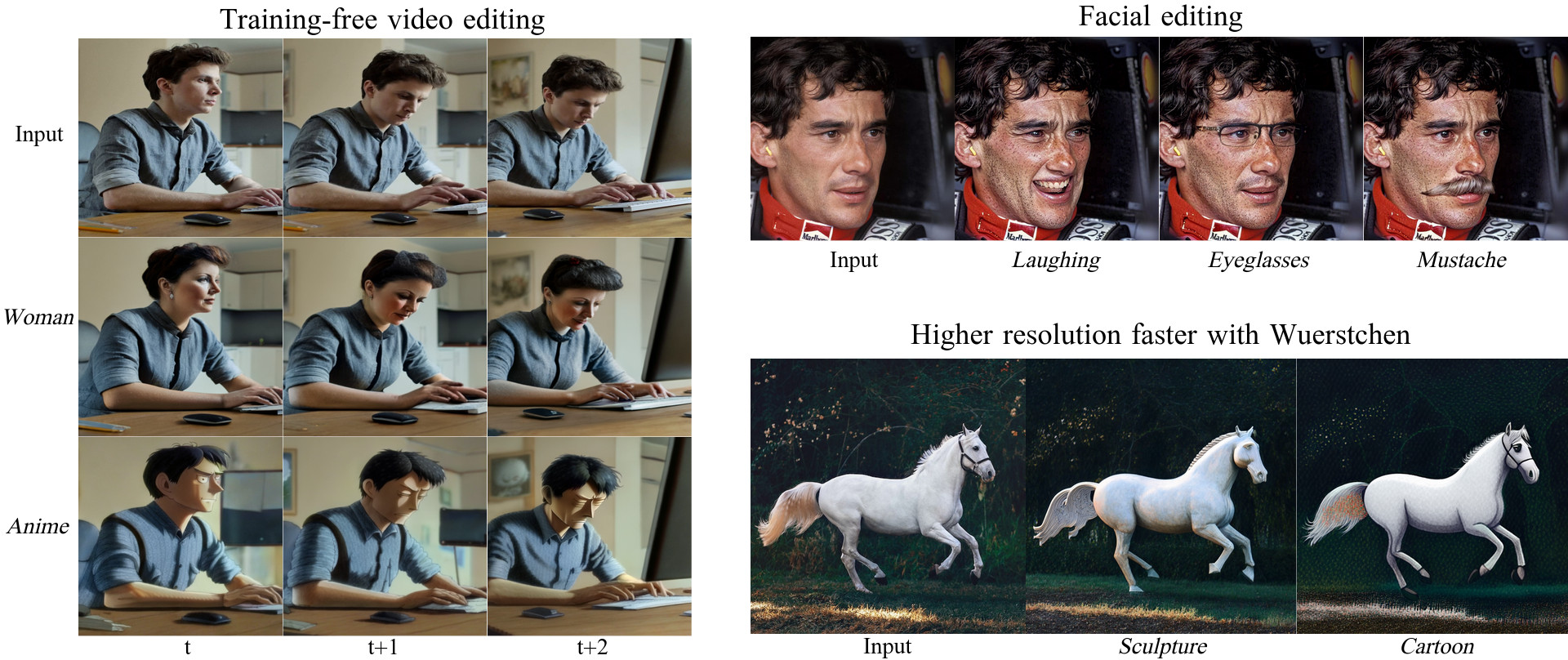}
  \caption{LASPA shows promise for future work in video editing, facial editing, and efficient editing.}
\label{fig:vers}
\end{figure}
\section{Future work and Limitations}
We find that spatial latent alignment is versatile and shows promise for future work in video editing, facial editing, and more efficient image editing. This is illustrated in Figure~\ref{fig:vers}. We find that when applied to frames of a video, our method achieves frame consistency without temporal information. With stricter alignment, our method leads to high-quality facial editing results. We note that the writing on the clothing is unchanged by our method. Finally, our method is not restricted to stable diffusion. We use Wuerstchen~\cite{pernias2023wuerstchen} to showcase $1024\times1024$ resolution results in only 3 seconds.\par
\begin{figure}[t]
  \centering
 \includegraphics[width=\textwidth]{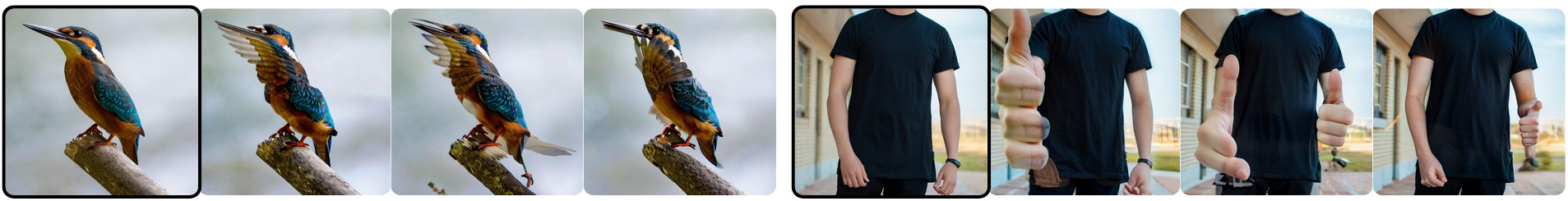}
  \caption{Example failure cases of our method. Input images are outlined. The edits are: spreading wings, and giving a thumbs up. Three random results are generated for each input image.}
\label{fig:failure}
\end{figure}

In terms of limitations, similar to other image editing methods, we find that the results can be improved by tuning parameters based on the kind of editing and by selecting better seeds. Another limitation is the difficulty of achieving some large pose changes. We show some failure cases of this kind in Figure~\ref{fig:failure}. Evidently, some kinds of edits are more difficult to achieve given only a single image and require tools such as MasaCtrl\cite{cao2023masactrl}.

\section{Conclusion}
This paper introduces LASPA, a novel framework for single-image editing using pre-trained text-to-image diffusion models. LASPA achieves significant improvements in both editing speed and quality, consistently generating edits in under 6 seconds while surpassing the quality of previous methods. This efficiency stems from LASPA's utilization of the spatial latent for editing, eliminating the need for computationally expensive fine-tuning or storing separate models for each edit.\par
LASPA offers three distinct approaches to latent spatial alignment for effectively preserving input image details throughout the editing process. These approaches involve aligning the input at each step, aligning the predicted initial latent state, or directly aligning the denoising error. Additionally, LASPA demonstrates the semantic value of its aligned latents by utilizing the pre-trained model's attention maps to achieve accurate background modifications.\par%This capability allows LASPA to produce high-quality edits even when substantial background changes are required, including adjustments to shadows and ambient occlusion effects.\par
LASPA's key strengths lie in its exceptional speed, minimal computational requirements, and competitive editing quality. This paves the way for the development of user-friendly and customizable image editing applications, particularly in scenarios where traditional approaches are impractical due to their computational demands. We believe LASPA represents a significant step forward in diffusion-based image editing, offering a powerful tool for both researchers and developers seeking to create fast, high-quality, and accessible editing experiences. %Unlike existing methods that often necessitate lengthy training and storage of individual models per edit, LASPA operates solely during inference, making it highly accessible and efficient. This paves the way for the development of user-friendly and customizable image editing applications, particularly in scenarios where traditional approaches are impractical due to their computational demands.\par


\bibliographystyle{splncs04}
\bibliography{main}

\clearpage
\setcounter{section}{0}
\renewcommand\thesection{\Alph{section}}

\section{Supplementary materials}

\begin{figure}
  \centering
 \includegraphics[width=0.8\textwidth]{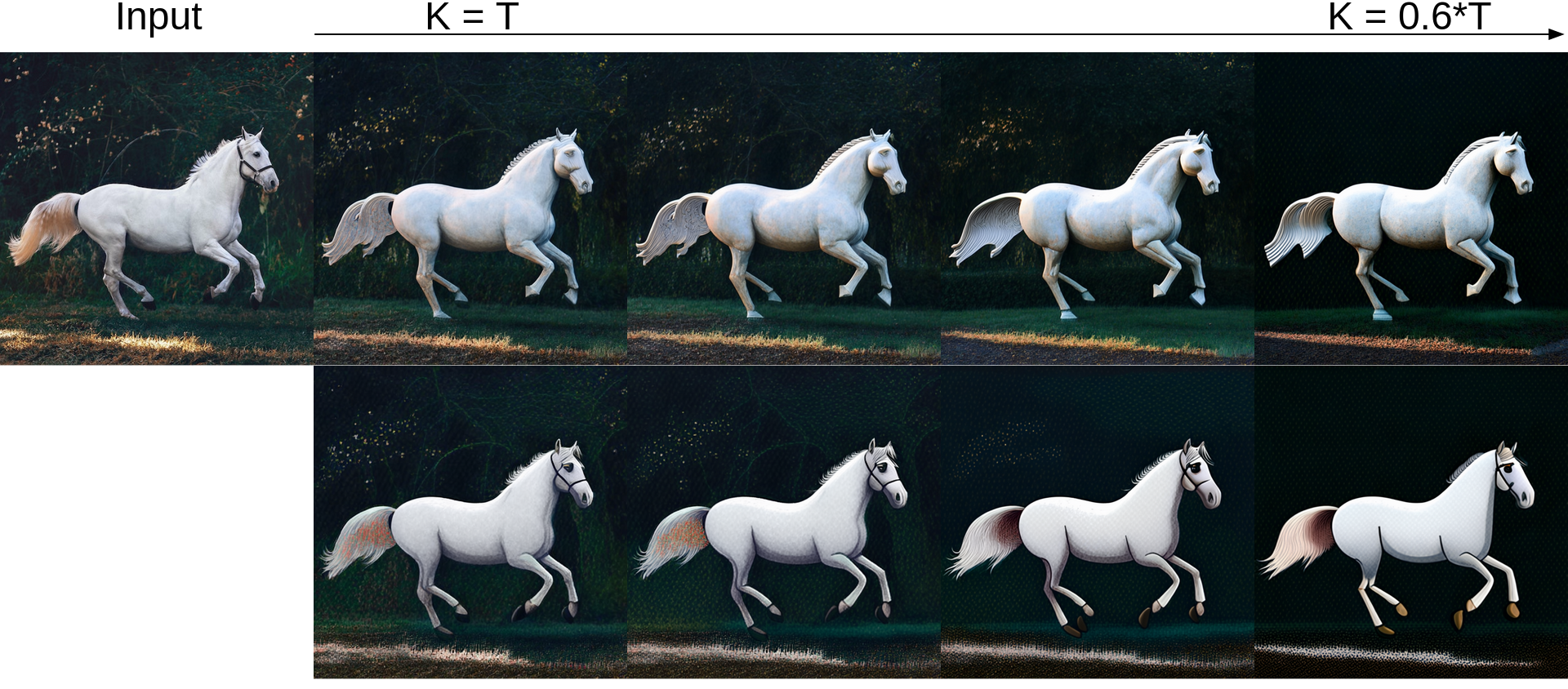}
  \caption{Results with Wuerstchen. The images are $1024\times1024$ and each editing result is obtained in around 3 seconds. $K$ controls the interval of alignment. Shorter alignment intervals lead to increased editing strength but less identity preservation. The edits are sculpture and cartoon}
    \label{fig:wurst}
\end{figure}
\subsection{LASPA for videos and mobile diffusion}
Our method has a strong advantage over prior methods in terms of speed. This renders our method more suitable for processing many frames and for results on mobile devices.\par
We show several clips on the webpage to showcase the performance on videos. We convert the input clips to frames, and we process each frame using our method. Even without the temporal information, our method shows promising results for future work. We test several realistic edits such as object replacement, as well as artistic edits. By fixing the unrestricted random latent for all frames, some continuity is achieved as seen in the attached videos.\newline
Additionally, we demonstrate how the speed can be further increased by using text-to-image diffusion models that are faster than stable diffusion. We use Wuerstchen~\cite{pernias2023wuerstchen} which can generate $1024\times1024$ images in 3 seconds by using smaller inner representations. While the Wuerstchen uses DDPM sampling, we can still edit using input alignment. We showcase our results with high resolution images using different choices of alignment interval end $K$ in Figure~\ref{fig:wurst}.\par

\begin{figure}[t]
  \centering
 \includegraphics[width=0.8\textwidth]{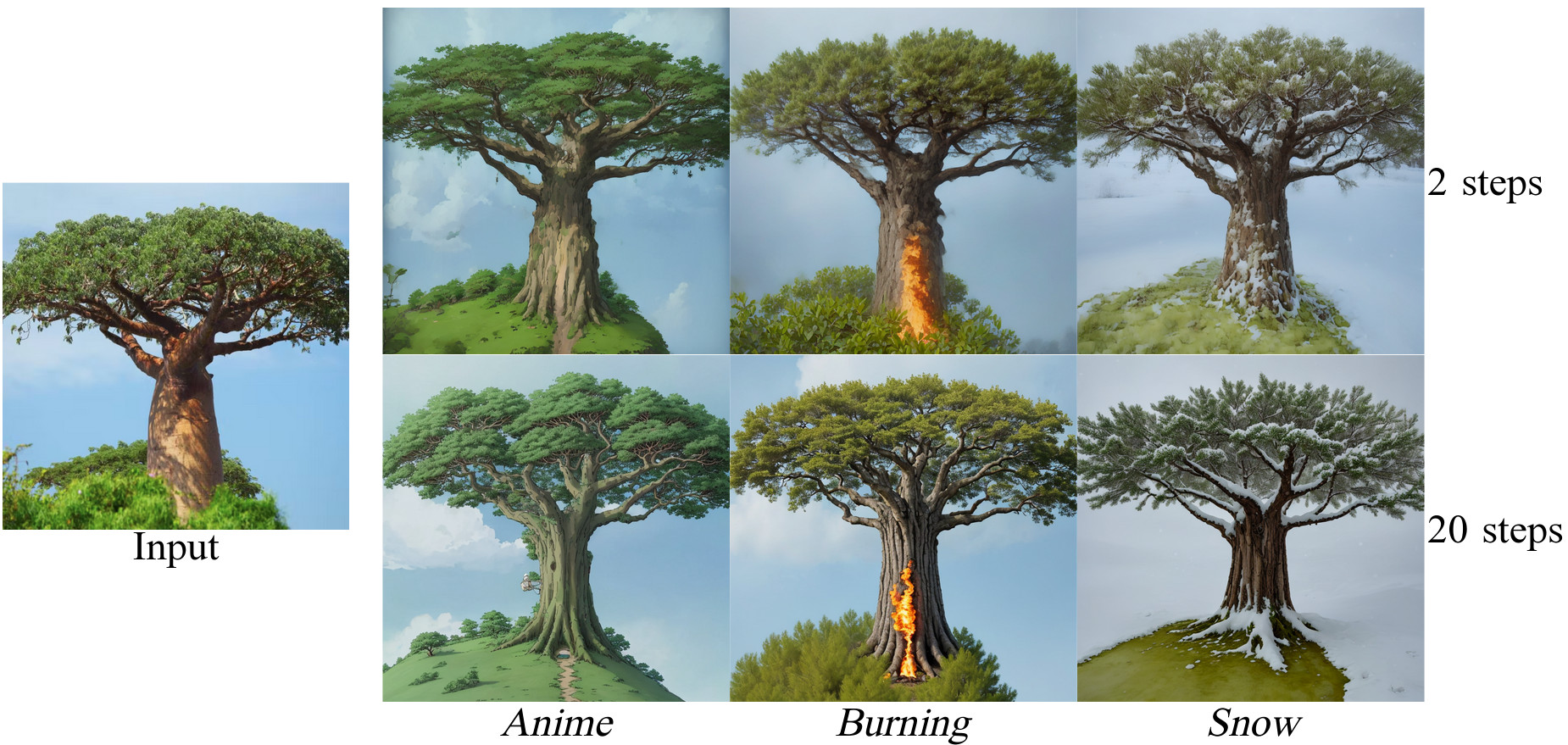}
  \caption{LASPA can edit using LCMs in as few as two steps.}
    \label{fig:lcm}
\end{figure}

\subsection{LASPA for LCM models}
Recently, Latent Consistency Models (LCMs) for image synthesis~\cite{lcm} achieved high quality generation with fewer steps than pretrained stable diffusion. We show how our method naturally extends to LCM models. We use prediction of $x_0$ alignment to produce the results in Figure~\ref{fig:lcm}. All results were produced using the same parameters.

\subsection{Facial editing with LASPA}
\begin{figure}
  \centering
 \includegraphics[width=0.8\textwidth]{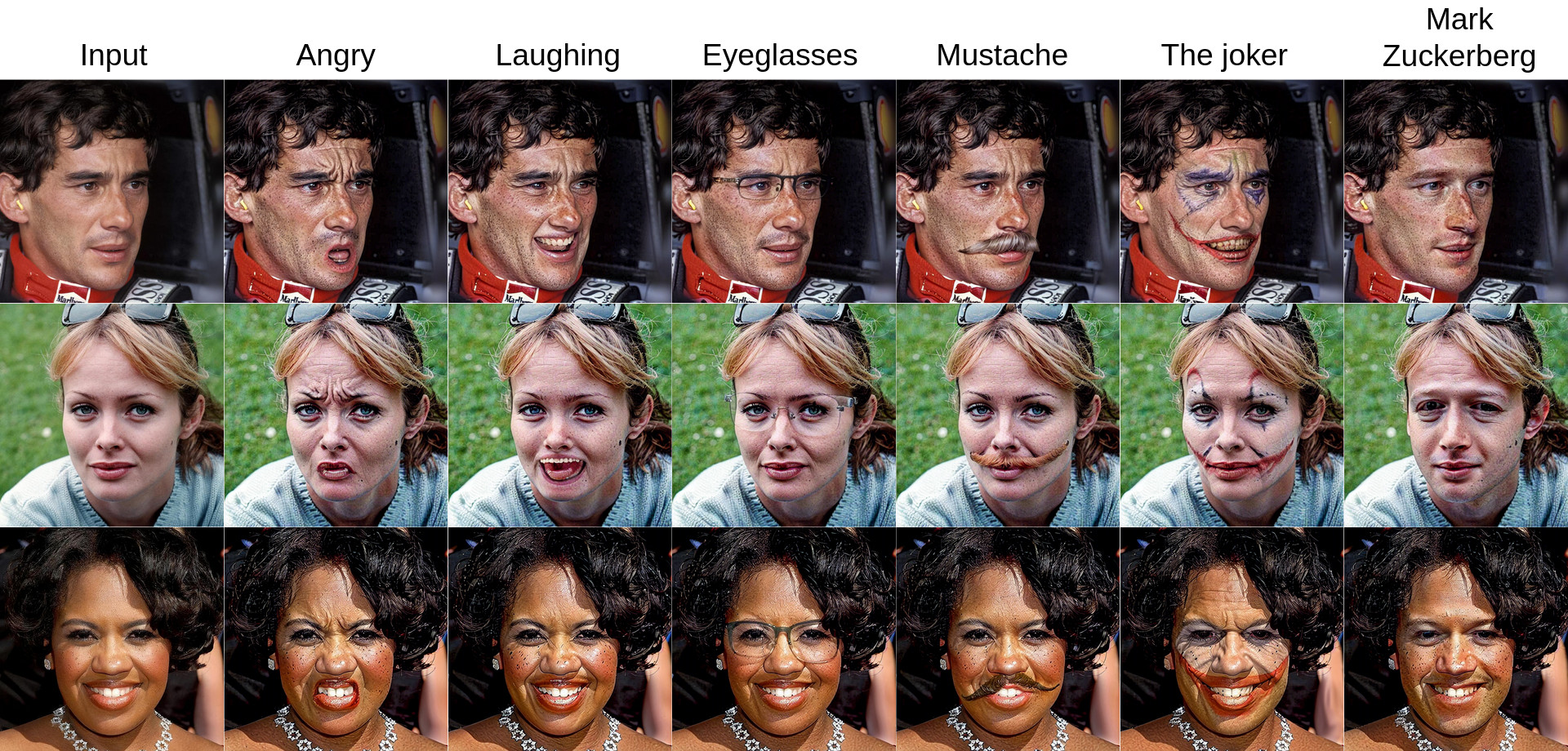}
  \caption{Facial editing results with our method. We observe that our method preserves details such as text on the clothing, accessories, and unique facial features.}
    \label{fig:facial}
\end{figure}

As mentioned in the paper, accurate facial editing is possible with stricter parameters (lower scale, longer alignment interval, and higher interpolation coefficient). All the results in Figure~\ref{fig:facial} are produced with the same settings.\newline
We can see that image preservation is very high. Letters in the backgrond, accessories, and unique facial features are preserved with high accuracy. This is incredibly difficult to achieve with other generative models such as GANs.

\begin{figure}
  \centering
 \includegraphics[width=0.8\textwidth]{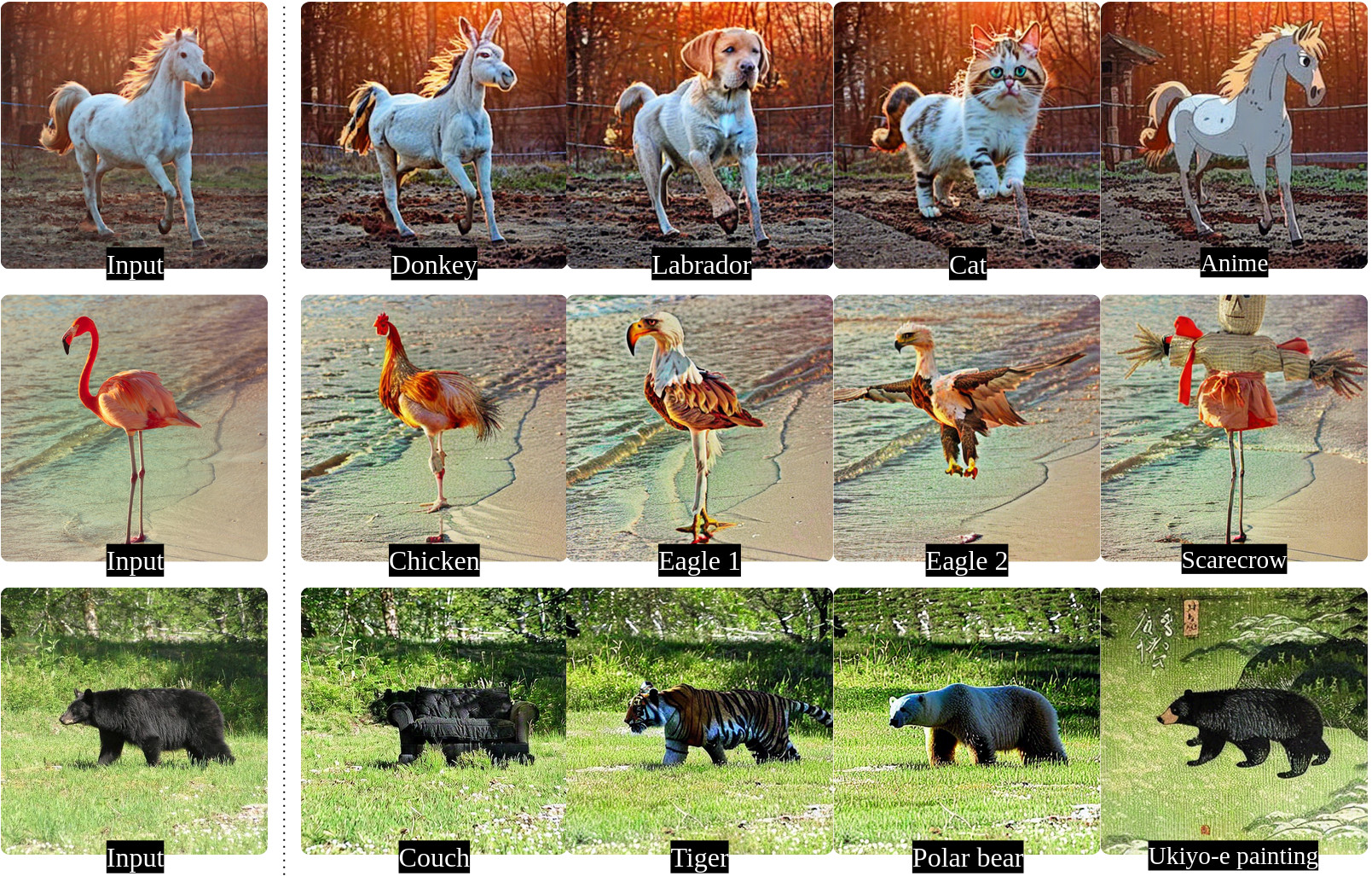}
  \caption{Extended results on TEdBench samples.}
    \label{fig:tedbench}
\end{figure}
\subsection{Additional results}
\begin{figure}
  \centering
 \includegraphics[width=\textwidth]{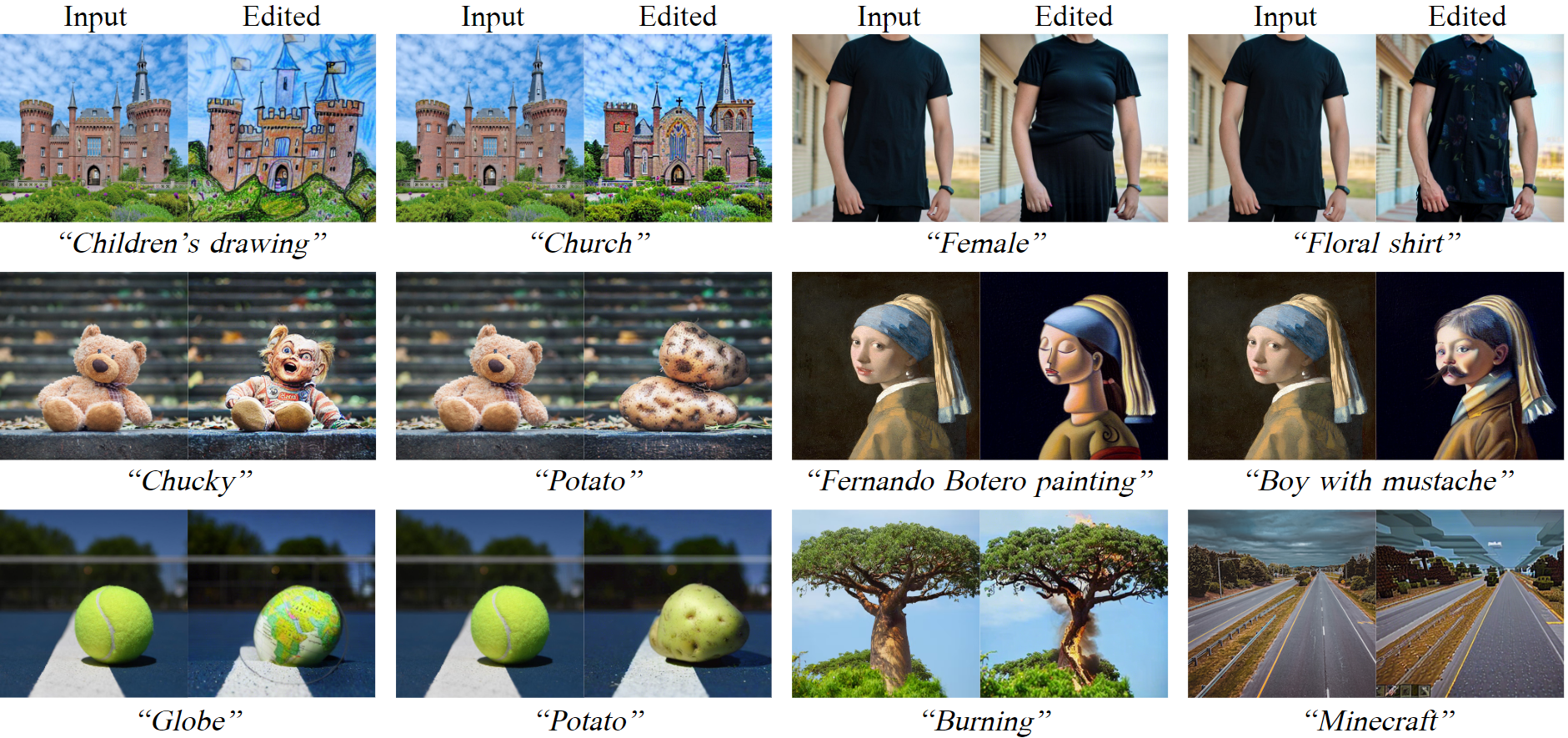}
  \caption{Extended results on a variety of images and edits.}
    \label{fig:add}
\end{figure}

We present additional results to showcase our method.\newline
Some animal swapping results from TEdBench show interesting results. The results are shown in Figure~\ref{fig:tedbench}. We can see the quality of object replacement edits, even when changes to the outline of the object are required. In the middle row, we highlight the eagle example where different degrees of editing can be achieved. Eagle1 maintains the outline of the reference image exactly, leading to an eagle image that is more accurate but less realistic. Eagle2 shows a different possible result where the outline is not followed strictly, but the resulting image is more natural.\newline
Additionally, we present additional results on a variety of images and edits in Figure~\ref{fig:add}. 

\subsection{Additional visualizations of alignment}
\begin{figure}
  \centering
 \includegraphics[width=0.8\textwidth]{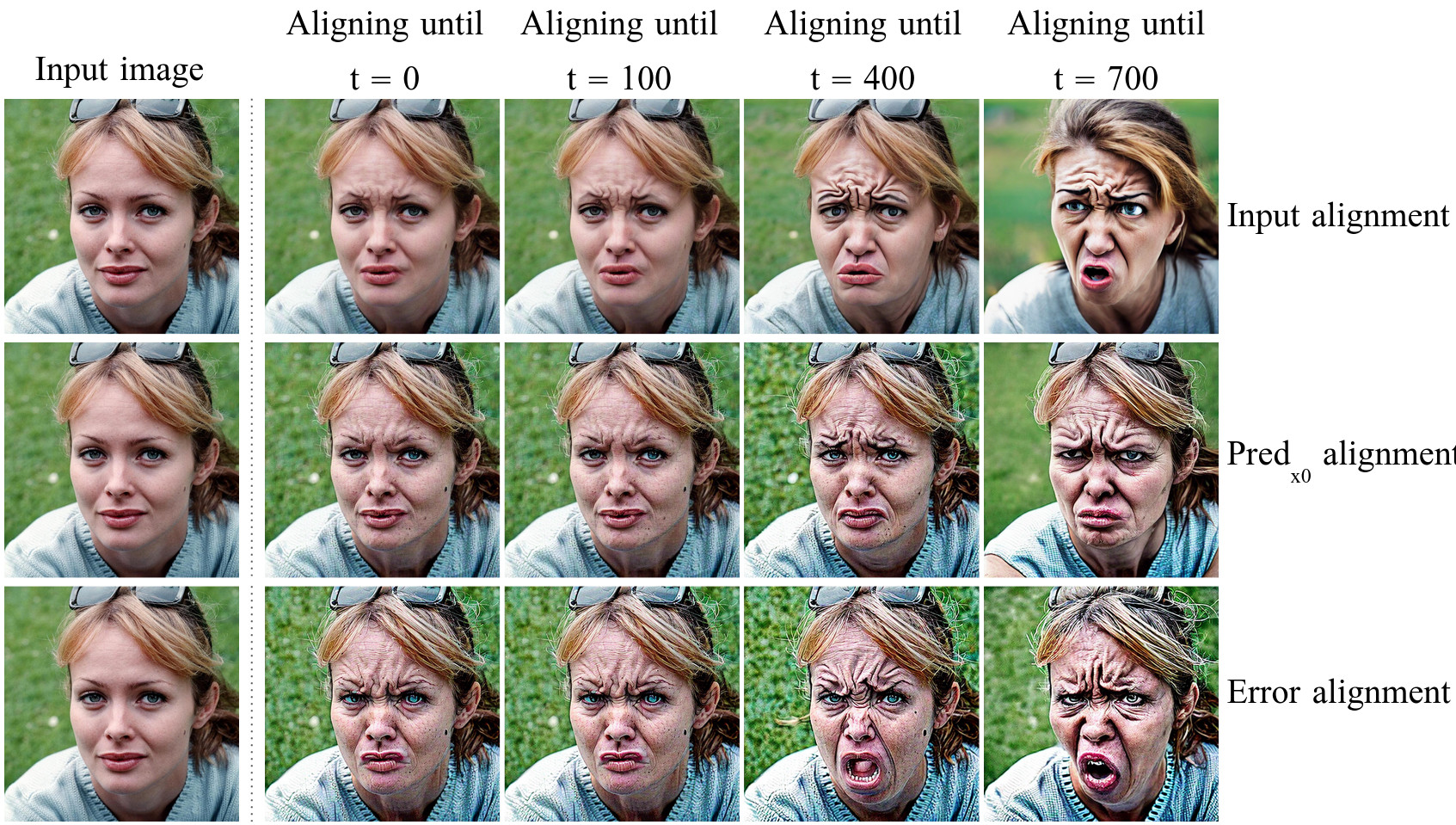}
  \caption{Extended visualization of our latent alignment. The editing prompt is ``a photo of an angry woman".}
    \label{fig:intervis}
\end{figure}

In the paper, we visualize the intermediate output when using input alignment (Figure 3, page 6). Here, we show results with $predx_0$ and $\epsilon_{\theta}$ alignment. All of the results in Figure~\ref{fig:intervis}, were obtained with stable diffusion 2.1 on resolution $768\times768$.\newline
We can see that all three approaches lead to results that are faithful to both the input image and the prompt. Furthermore, the intermediate visualizations show that they are following a smooth and reasonable path of editing. Input alignment if applied in all steps (until t=0) leads to strong preservation of input image details. On the other hand, $pred_{x0}$ alignment leads to stronger textual influence than input alignment. Error alignment while showing strong textual influence leads to lower image quality than input and $pred_{x0}$ alignment.%Both $\epsilon_{\theta}$ and $predx_0$ alignments, by design and if $\beta=1$, lead to starting the diffusion path with predicting the final result to be almost exactly the input image. This is not the case in input alignment with the stochastic inversion, since in early steps the stochastic inversion has a high weight for the noise and a low weight for the input image.

\subsection{Additional comparisons against SINE}
\begin{figure}[t]
  \centering
 \includegraphics[width=0.8\textwidth]{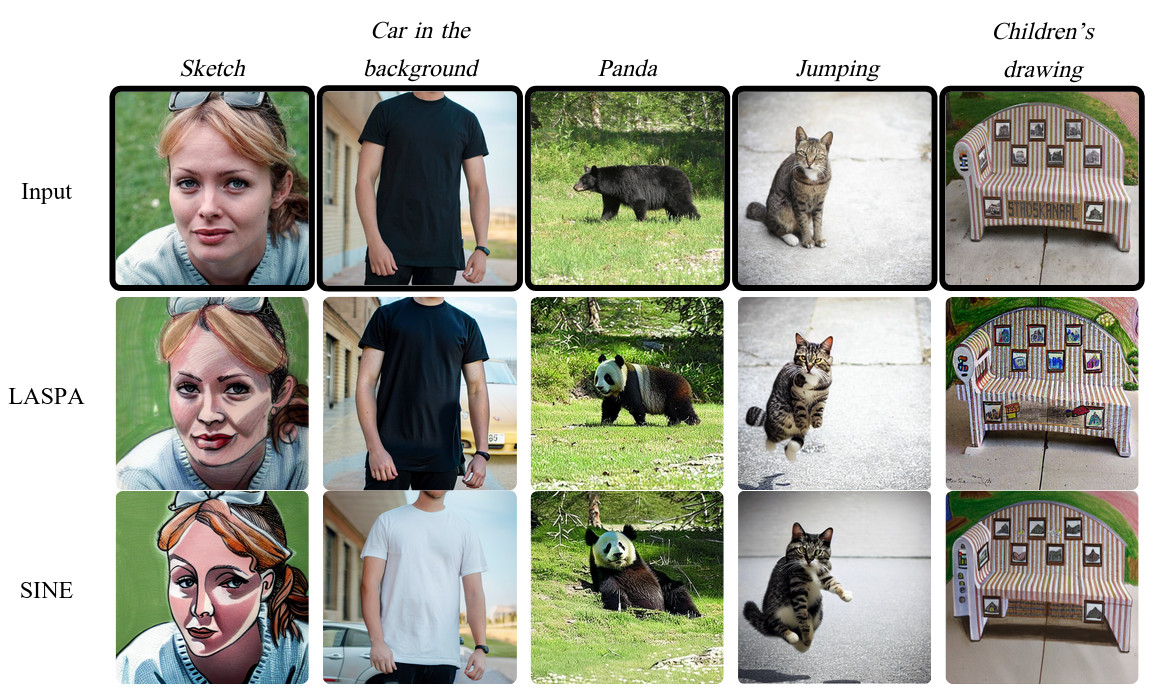}
  \caption{SINE struggles to maintain some reference image details}
    \label{fig:addagnstsine}
\end{figure}

In Figure~\ref{fig:addagnstsine} we present additional results compapring LASPA to SINE. In most cases, SINE makes irrelevant changes to the shapes or colors of the reference image. LASPA, on the other hand, reflects the required textual edits with minimal changes to the input image. We believe that this is a direct result of using the spatial latent of the reference image to align. Shapes and colors of most irrelevant details are well preserved by LASPA. As we discuss in the following section, SINE finetunes an entire stable diffusion network with a modified prompt to preserve the input image. This proves to be problematic in some cases as reflected in their results.
\subsection{Contrast with SINE}
While Imagic optimizes a textual embedding as well as the network, SINE is more comparable to our method as it does not directly optimize the textual embedding. Instead, borrowing from previous works, SINE selects a special uncommon word (namely ``sks") and finetunes the network to reproduce the input image given the adjusted prompt. So given an image of a castle, SINE will finetune a stable diffusion network with the prompt ``a photo of a sks castle". The finetuned network serves only a single goal, which is reproducing the input image. That is the major difference between how SINE and our method preserve the input image.\newline
The methods also differ in how edits are achieved. For SINE, two classifier-free guidance results are interpolated and they come from two different networks (the finetuned network and pretrained stable diffusion). In our case, we align latents of the pretrained network using \textit{computed} reference latents.\newline
Nonetheless, there are several shared ideas. Similar to our method, pretrained stable diffusion is utilized in SINE without modification as the source of word influence. Finetuning is required per image, but editing on the same image can be done at inference time with the same finetuned network. In addition, $K$ and $v$ in SINE achieve very similar effects to LASPA's $K$ and $\beta$. In a sense, SINE does latent alignment as well, albeit using a network finetuned with a special textual token. In general, SINE and LASPA share strengths and pitfalls. Both are significantly more faithful to the input image than Imagic. On the other hand, they both suffer from low quality results on large pose changes or with certain seeds.\newline
However, with regards to performance LAPSA is significantly faster and less complex to run. SINE requires 15 minutes of finetuning per image, storing a model per image, and running two models simultaneously at inference time. For datasets like TEdBench, the total size of the models is hundreds of Gigabytes and the total time is several hours. LASPA, on the other hand, does not require storing additional models besides the pretrained stable diffusion and can perform any edit in 6 seconds.

\end{document}